\documentclass[10pt,twocolumn,letterpaper]{article}

\usepackage[pagenumbers]{cvpr} 

\usepackage[dvipsnames]{xcolor}

\definecolor{cvprblue}{rgb}{0.21,0.49,0.74}
\usepackage[pagebackref,breaklinks,colorlinks,citecolor=cvprblue]{hyperref}

\usepackage[noend]{algpseudocode}

\usepackage{algorithmicx,algorithm}

\usepackage{bbding}
\usepackage{pifont}
\usepackage{color}
\usepackage{amsmath}
\usepackage{amssymb}

\begin{document}

\title{Balanced Multi-modal Federated Learning via Cross-Modal Infiltration}

\author{%
   Yunfeng~Fan$^1$,
   Wenchao~Xu$^1$,
   Haozhao~Wang$^2$,
   Jiaqi~Zhu$^3$,
   and~Song~Guo$^1$\\
  \textsuperscript{1}Department of Computing, The Hong Kong Polytechnic University\\
  \textsuperscript{2}School of Computer Science and Technology, Huazhong University of Science and Technology\\
  \textsuperscript{3}School of Automation, Beijing Institute of Technology\\
  \texttt{yunfeng.fan@connect.polyu.hk}, \texttt{wenchao.xu@polyu.edu.hk}, \texttt{hz\_wang@hust.edu.cn}, \\
  \texttt{E1111838@u.nus.edu}, \texttt{songguo@cse.ust.hk}
}

\maketitle
\begin{abstract}

Federated learning (FL) underpins advancements in privacy-preserving distributed computing by collaboratively training neural networks without exposing clients’ raw data. Current FL paradigms primarily focus on unimodal data, while exploiting the knowledge from distributed multimodal data remains largely unexplored. Existing multimodal FL (MFL) solutions are mainly designed for statistical or modality heterogeneity from the input side, however, have yet to solve the fundamental issue, `modality imbalance’, in distributed conditions, which can lead to inadequate information exploitation and heterogeneous knowledge aggregation on different modalities.In this paper, we propose a novel Cross-Modal Infiltration Federated Learning (FedCMI) framework that effectively alleviates modality imbalance and knowledge heterogeneity via knowledge transfer from the global dominant modality. To avoid the loss of information in the weak modality due to merely imitating the behavior of dominant modality, we design the two-projector module to integrate the knowledge from dominant modality while still promoting the local feature exploitation of weak modality. In addition, we introduce a class-wise temperature adaptation scheme to achieve fair performance across different classes. Extensive experiments over popular datasets are conducted and give us a gratifying confirmation of the proposed framework for fully exploring the information of each modality in MFL.


\end{abstract}   
\section{Introduction}
\label{sec:intro}
Federated learning (FL) has emerged as a predominant framework for collaboratively training a shared model while keeping the users' raw data locally, which is widely applied in broad intelligent applications, such as smart healthcare \cite{pfitzner2021federated}, financial services \cite{byrd2020differentially} and smart city \cite{zheng2022applications}. The majority of previous endeavors have been dedicated to exploring the common priors from users' unimodal data.
However, the growing penetration of diverse sensor technologies in smart devices has spurred a surge of interest in multimodal FL (MFL), which can jointly utilize the distributed multimodal data among different users \cite{chen2022towards, zheng2023autofed, liu2020federated, xiong2022unified}.   

\begin{figure*}[t]
    \centering
    \begin{subfigure}{0.325\linewidth}
        \includegraphics[width=0.9\linewidth]{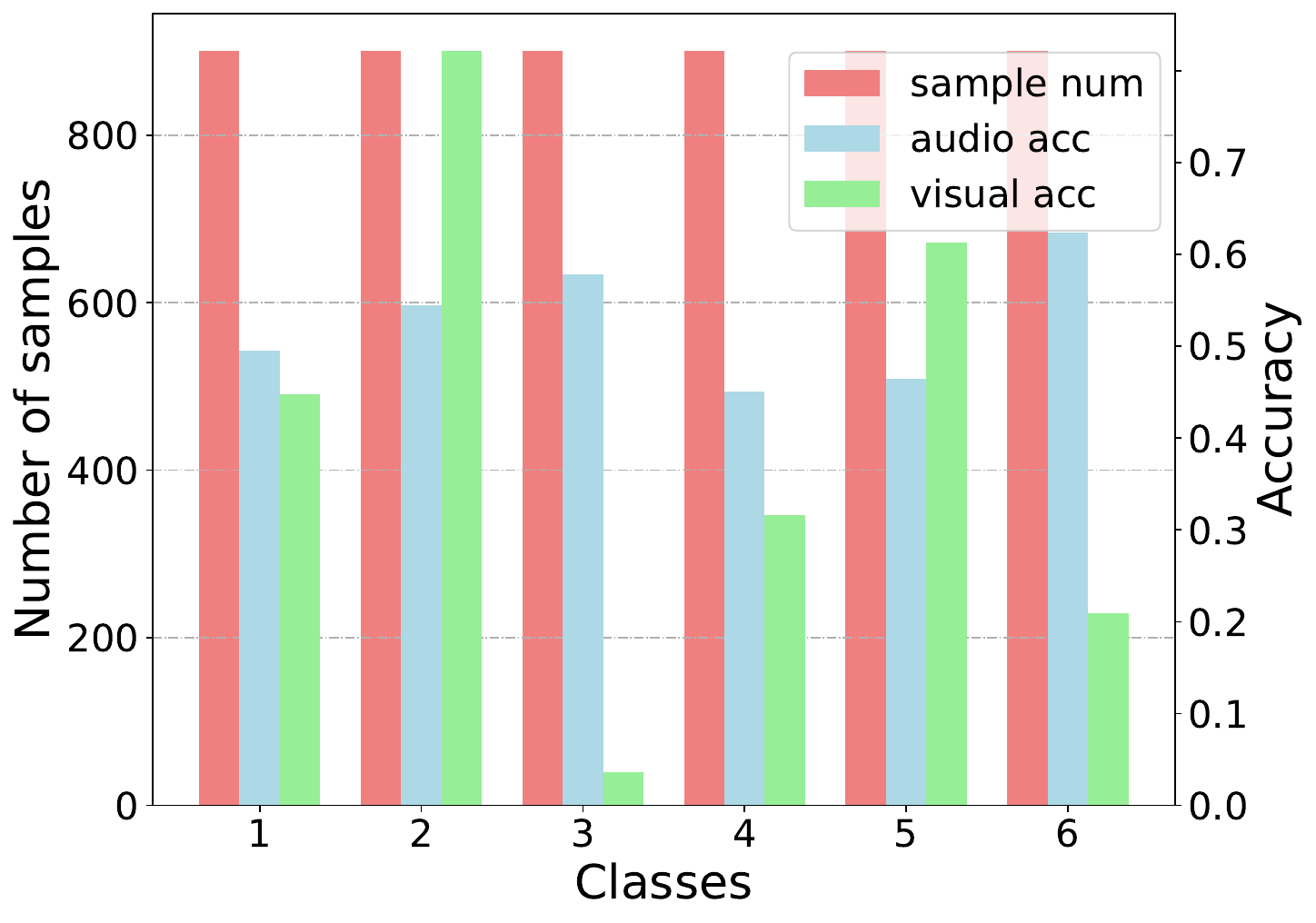}
        \caption{Client 1}
        \label{fig:sub_figure1}
    \end{subfigure}
    \hfill
    \begin{subfigure}{0.325\linewidth}
        \includegraphics[width=0.9\linewidth]{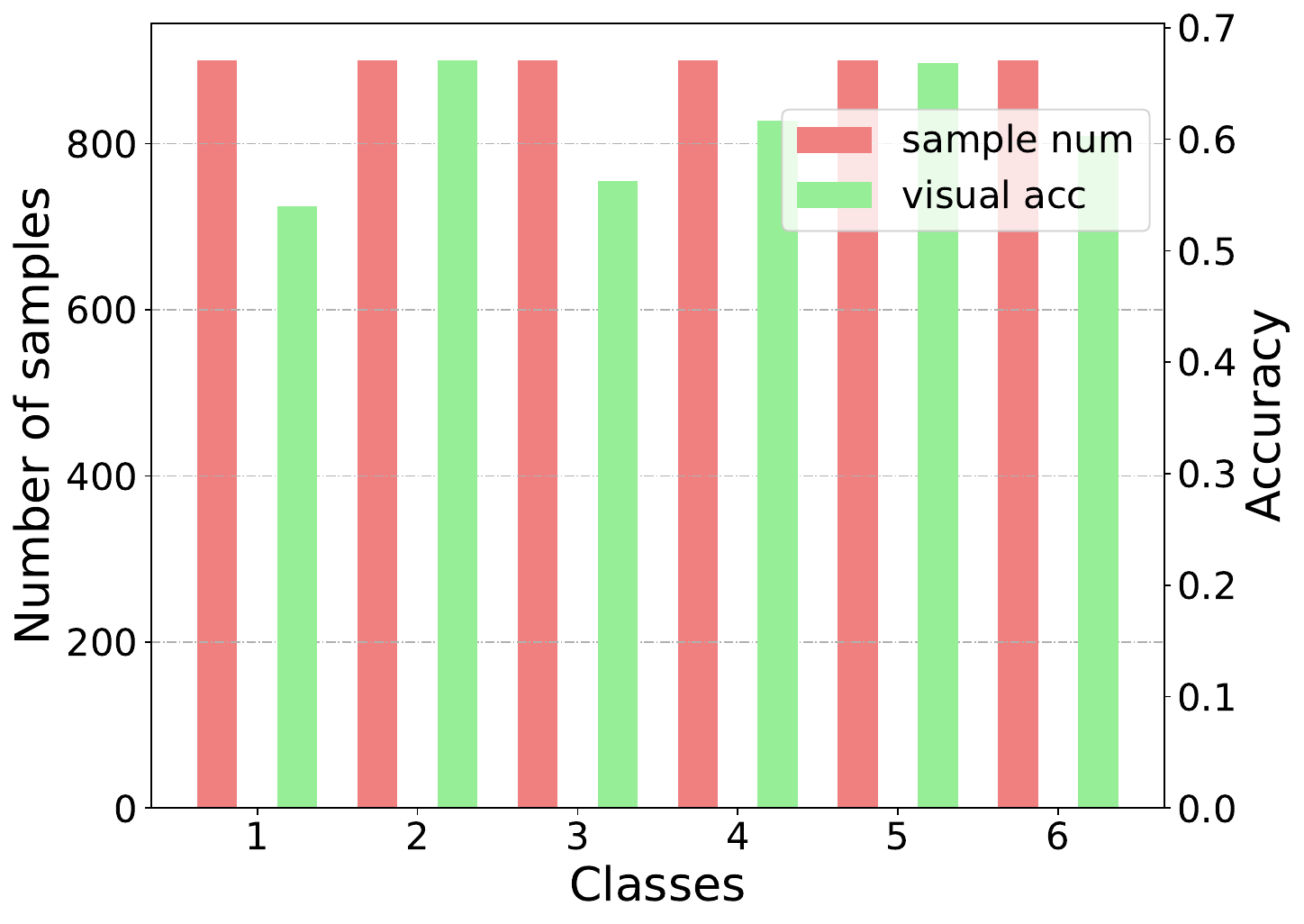}
        \caption{Client 2}
        \label{fig:sub_figure2}
    \end{subfigure}
    \hfill
    \begin{subfigure}{0.325\linewidth}
        \includegraphics[width=0.9\linewidth]{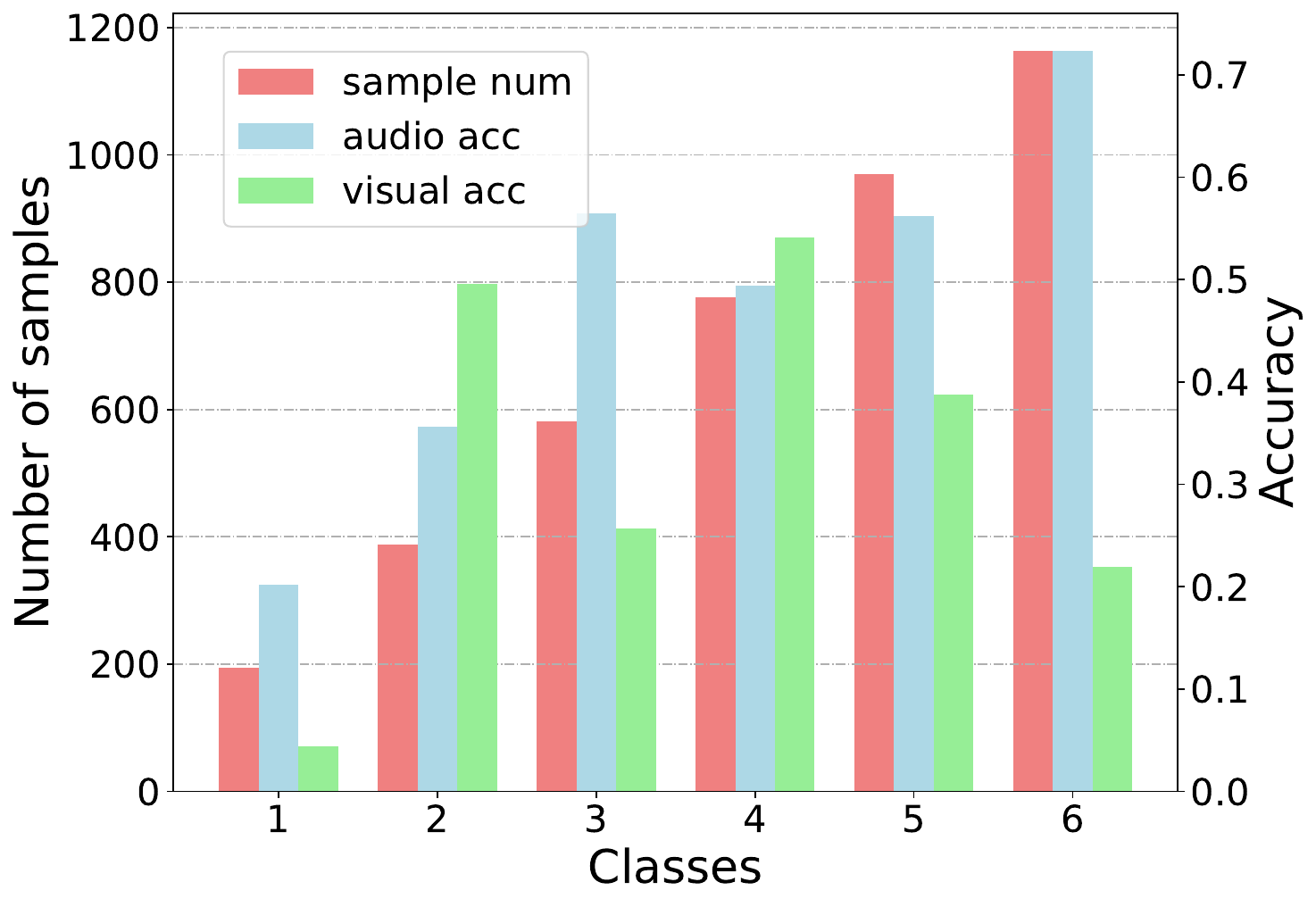}
        \caption{Client 3}
        \label{fig:sub_figure3}
    \end{subfigure}
    \caption{The performance of each class for different modalities on CREMA-D with vanilla local training strategy. Modality imbalance behaves differently in clients with different modalities of data and diverse data distributions (client 1 and client 3 possess both audio and visual data with different distributions. Client 2 contains only visual data with the same distribution as client 1).}
    \label{fig: motivation}
\end{figure*}

Literature on the topic of MFL has proposed to address the statistical \cite{yu2023multimodal,xiong2022unified} or modality heterogeneity \cite{zheng2023autofed, chen2022fedmsplit} from the input side (noted as \textit{input heterogeneity}), while a critical issue inherent to multimodal learning, i.e., \textit{modality imbalance} which occurs during the learning process \cite{peng2022balanced, wu2022characterizing, xiao2020audiovisual}, has been overlooked in existing studies. 
Modality imbalance indicates the inconsistent learning speeds of different modalities in multimodal learning. As discussed in \cite{wang2020makes,peng2022balanced}, the dominant modality that learns faster would hinder the learning process of slower modalities, leading to insufficient exploration for multimodal data. We point out that in addition to the underutilization of information, there are also two major challenges in MFL when jointly considering modality imbalance and input heterogeneity:
\\
$\bullet$ \textit{Challenge 1. Heterogeneous modality inhibition across clients.} 
In MFL, we find that the inhibition phenomenon on weak modality greatly differs among clients, as shown in \cref{fig: motivation}. The inhibition on visual modality (weak) is remarkable in client 1 and client 3 but varies between each other because of divergent data distributions, while client 2, without inhibition from audio data, performs much better on visual modality even though there is statistical homogeneity compared with client 1.
\\
$\bullet$ \textit{Challenge 2. Inconsistent class-wise performance.} As demonstrated in \cref{fig:sub_figure1,fig:sub_figure3}, the visual modality is not only severely inhibited but also suffers from great performance divergence on different classes. The differences in classes do not follow changes in data distributions (a huge gap between class 2 and class 3 in client 1 with the same sample number, while class 6 performs worse even with more samples), which makes the class-wise performance across clients inconsistent and unpredictable.

Many prior initiatives have separately considered the problems of modality imbalance and input heterogeneity. However, the approaches for input heterogeneity \cite{yu2023multimodal,zheng2023autofed} fail to address the underutilization of information while the methods for modality imbalance \cite{peng2022balanced,fan2023pmr} are generally designed for centralized settings and cannot be directly applied in distributed scenarios due to the substantial heterogeneity among clients. Therefore, we aim to tackle the two mentioned challenges derived from jointly considering modality imbalance and input heterogeneity.

In this paper, we propose a novel \textbf{C}ross-\textbf{M}odal \textbf{I}nfiltration \textbf{Fed}erated learning (FedCMI) framework to integrate the knowledge from the global dominant modality to local weak modality while still maintain the information exploitation for each modality. 
Specifically, the core idea of our method is that the global dominant modality acquires sufficient knowledge from all clients via iterative aggregations, which can be applied as an excellent teacher for the local weak modality since there exists shared knowledge between them \cite{wang2021multimodal, tian2022multimodal, yang2022disentangled}.
However as stated in \cite{xue2022modality}, knowledge distillation (KD) is not a universal solution for cross-modal knowledge transfer as modality-specific knowledge still exists. 
%
%
Merely imitating the behavioral attributes of the dominant modality may lead to the loss of the information from weak modality.
%
Therefore, we design a two-projector architecture as shown in \cref{fig: framework of CMI}. The infiltration projector receives the distillation signal to integrate the knowledge from dominant modality and the self-projector is responsible for information exploration of its own modality. The two-projector design allows a unimodal network to preserve the multimodal knowledge, like \textit{one modality being infiltrated by another}.  
The principles of why our method can simultaneously solve the two challenges are twofold. First, the global dominant modality can not only boost the learning of weak modality but also provide consistent knowledge for participating clients, which helps to alleviate the heterogeneous modality inhibition. Second, we further propose a class-wise temperature adaptation scheme to mitigate the remarkable performance divergence on different classes of the weak modality.
To summarize, the contributions of the paper are as follows:
\begin{itemize}
\item We propose a novel MFL framework, namely FedCMI, to resolve the challenges of jointly considering modality imbalance and input heterogeneity.
\item The two-projector design in FedCMI stimulates the weak modality by integrating the knowledge from the global dominant modality while still maintains the information exploitation of each modality. We further adopt a class-wise temperature adaptation scheme to alleviate the knowledge bias through different classes. 
\item We conduct comprehensive experiments and demonstrate that FedCMI can achieve considerable improvements over baselines on global model in various scenarios.
\end{itemize}


\section{Related Work}
\label{sec: related}
\subsection{Unimodal Federated Learning}
Federated learning (FL) \cite{mcmahan2017communication} works to jointly train a global model with a large number of clients while preserving privacy. To tackle the statistical heterogeneity in FL, FedProx \cite{li2020federated} adds a proximal term to the objective that helps to improve the stability. FedProto \cite{tan2022fedproto} shares the abstract class prototypes instead of the gradients between server and clients to regularize the training of local models.
FedNH \cite{dai2023tackling} improves the generalization of the global model by distributing class prototypes uniformly in the latent space to solve the class imbalance setting. 
However, current methods mainly focus on unimodal settings, which makes it hard to satisfy the increasing demand for multimodal scenarios.

\subsection{Multimodal Federated Learning}
Only limited attempts have been made to solve multimodal tasks in FL (MFL). 
FedIoT \cite{zhao2022multimodal} is a multimodal FedAvg \cite{mcmahan2017communication} algorithm to extract correlated representations from local autoencoders.
%
FedMSplit \cite{chen2022fedmsplit} focuses on modality heterogeneity in MFL. It splits local models into several components and aggregates them by the correlations amongst multimodal clients according to a dynamic and multi-view graph structure. 
CreamFL \cite{yu2023multimodal} comprehensively takes into account statistical heterogeneity, model heterogeneity and task heterogeneity in MFL, and uses knowledge distillation \cite{hinton2015distilling} with contrastive learning \cite{chen2020simple} via a public dataset.
%
Although the literature considers various cases in MFL, the modality imbalance, which is vital in multimodal learning, has been ignored. In this paper, we mainly try to tackle the challenges of combining modality imbalance and input heterogeneity in MFL. 

\subsection{Imbalanced Multimodal Learning}
Modality imbalance \cite{wang2020makes} refers to the learning paces of different modalities in multimodal learning. The reason for such a problem is that modalities keep inhibiting each other, resulting in inadequate information exploitation for each modality, especially for the weak one. 
To tackle this problem, MSLR \cite{yao2022modality} customizes different learning rates for different modalities. 
Peng \etal \cite{peng2022balanced} propose OGM-GE to modulate the training pace of the dominant modality while monitoring its learning speed. 
Fan \etal \cite{fan2023pmr} find that the updating gradient directions can be disturbed by other modalities and propose PMR by introducing prototypes for gradient calibration. 
%
%
These methods are effective in centralized settings but cannot work well in distributed scenarios because of the heterogeneity among clients and their essential need for massive computing resources and data, which are usually unavailable in FL. 
They also ignore the potential of using other modalities to improve the performance of the weak modality.

\section{Method}
\label{sec:label}
\begin{figure*}[t]
  \centering
  \includegraphics[width=0.95\linewidth]{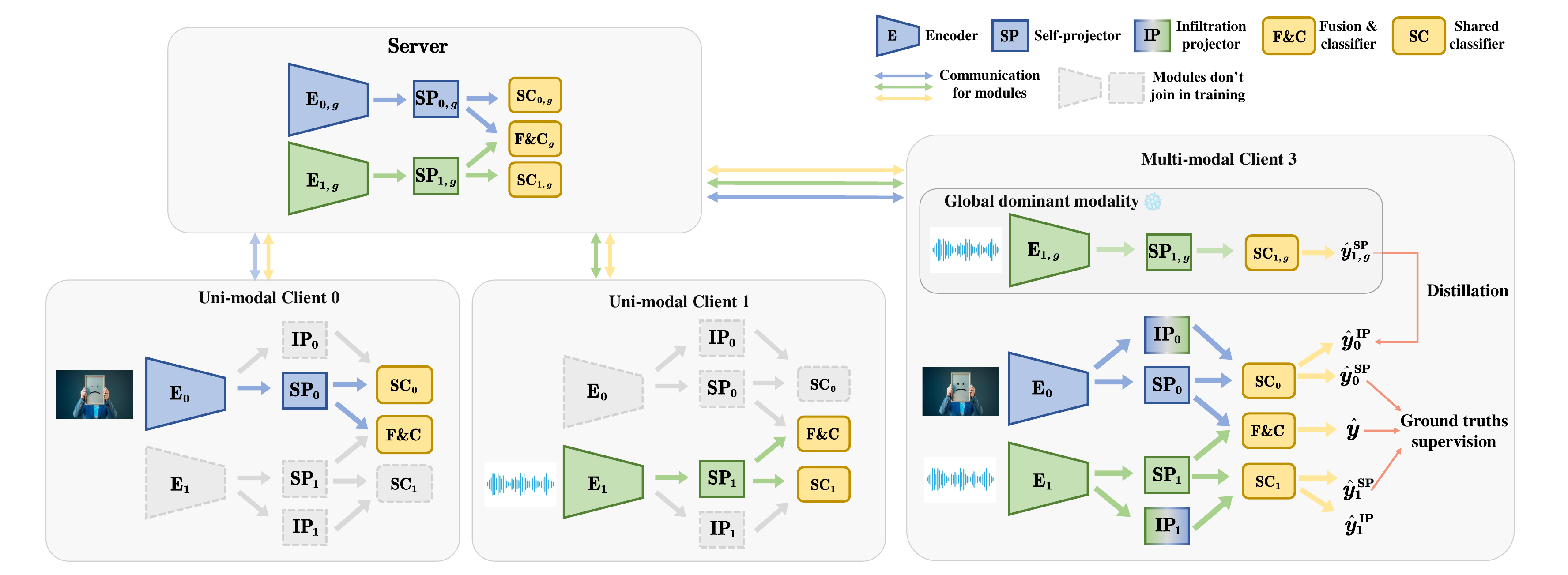}
  \caption{Overall workflow of the proposed framework. For multimodal clients, they exploit the information from each modality via ground truth supervision and also absorb the knowledge from the global dominant modality to alleviate heterogeneous modality inhibition. Unimodal clients only use local data to train corresponding modules. All updated modules except the infiltration projector participate in server-client communication.}
  \label{fig: framework of CMI}
\end{figure*}

\subsection{Problem Formulation}
\label{sec: mfl formulation}
Without loss of generality, we consider two input modalities, referred to as $m_0$ and $m_1$. We consider MFL with $N$ clients. 
Each client owns a local private dataset $D_i=\left\{ \boldsymbol{X}_{i}^{m_0},\boldsymbol{X}_{i}^{m_1},y \right\}, i\in \left\{ 1,2,\cdots ,N \right\}$ where the sample $\boldsymbol{X}_{i}$ and label $y$ are sampled from local distribution $\mathbb{P} _i$. Each modality in client $i$ has its own network backbone $\psi _{i}^{m_0},\psi _{i}^{m_1}$ with parameter $\boldsymbol{\theta }_{i}^{m_0},\boldsymbol{\theta }_{i}^{m_1}$ to extract features. Each client also owns $\mathbb{C} _i$, the combination of fusion module and joint classifier with parameter $\boldsymbol{\omega }_i$. 
The objective of MFL is to learn a global model $\left\{ \psi ^{m_0},\psi ^{m_1}; \mathbb{C} \right\} $ with parameter $\boldsymbol{\phi }=\left\{ \boldsymbol{\theta }^{m_0},\boldsymbol{\theta }^{m_1},\boldsymbol{\omega } \right\}$  that minimizes the empirical loss over the entire dataset $D=\left\{ D_1, D_2, \cdots , D_N \right\}$: 
\begin{equation}
\begin{aligned}
  \underset{\boldsymbol{\phi }}{\min}\mathcal{L} \left( \boldsymbol{\phi } \right) &:=\sum_{i=1}^N{\frac{\left| D_i \right|}{\left| D \right|}\mathcal{L} _i\left( \boldsymbol{\phi } \right)}
\\
\mathrm{where} \mathcal{L} _i\left( \boldsymbol{\phi } \right) &=\frac{1}{\left| D_i \right|}\sum_{k=1}^{\left| D_i \right|}{\mathcal{L} _{ct}\left( \boldsymbol{\phi };\boldsymbol{x}_{k}^{m_0},\boldsymbol{x}_{k}^{m_1},y_k \right)}
  \label{eq: global obj}
\end{aligned}
\end{equation}
where $\mathcal{L} _{ct}$ denotes an empirical loss for the classification task, \textit{e.g.} cross-entropy loss. 

In this paper, we consider MFL with two main settings: statistical and modality heterogeneity. Statistical heterogeneity (\textit{i.e.} non-IID data distributions across clients) means the local data are sampled from $N$ distinct distributions and have different sizes. Modality heterogeneity represents that the clients have different setups of sensors to perceive data (uni-$m_0$, uni-$m_1$ or both $m_0,m_1$). 

As discussed in \cite{fan2023pmr}, the gradient of the softmax logits output of summation fusion for cross-entropy loss with true label $y_i$ should be:
\begin{equation}
    \frac{\partial \mathcal{L} _{ct}}{\partial f\left( \boldsymbol{x}_i \right) _{y_i}}=\frac{e^{\hat{y}_{i}^{m_0}+\hat{y}_{i}^{m_1}}}{\sum\nolimits_{c=1}^C{e^{\hat{y}_{c}^{m_0}+\hat{y}_{c}^{m_1}}}}-1
    \label{eq: gradient}
\end{equation}
where $f\left( \boldsymbol{x}_i \right)$ means the function calculating the logit output for sample $\boldsymbol{x}_i$. $\hat{y}_{c}^{m}$ denotes the logit output for class $c$ with modality $m$. $C$ is the total number of classes. Suppose $m_0$ performs better, it contributes more to the gradient, diminishing the learning of $m_1$, which explains why the inhibited visual modality cannot achieve good performance even with more samples on class 6, as shown in \cref{fig:sub_figure3}.

\subsection{Cross-modal Infiltration}
\label{sec: cmi}
As mentioned above, the effect of modality imbalance varies among clients, which leads to inadequate information exploitation and significant knowledge heterogeneity, which impairs the generalization of the global model.
%
In order to alleviate the heterogeneous modality inhibition among clients, we aim to facilitate and balance the learning of weak modality via knowledge transfer from the global dominant modality. Therefore, we propose the FedCMI framework, as show in \cref{fig: framework of CMI}.

Each modality has its own encoder to extract features $z_{m_0},z_{m_1}\in \mathbb{R} ^{d_m}$. And then the extracted features are transformed by the self-projector (SP) and infiltration projector (IP) to $z_{m_0}^{SP},z_{m_0}^{IP},z_{m_1}^{SP}, z_{m_1}^{IP}\in \mathbb{R} ^{d_m}$. For modality $m_0$, on the one hand, $z_{m_0}^{SP}$ is fused with the features $z_{m_1}^{SP}$ and then passes through their joint classifier (C) to generate the logit $\hat{y}$ for classification. On the other hand, $z_{m_0}^{SP}$ and $z_{m_0}^{IP}$ pass through their shared classifier (SC) to output logits $\hat{y}_{m_0}^{SP}$ and $\hat{y}_{m_0}^{IP}$ for self-enhancement and cross-modal distillation respectively (following the similar pattern for $m_1$). We can simply acquire two cross-entropy losses $\mathcal{L} _{ce}$ and $\mathcal{L} _{ce}^{m_0}$ derived from $\hat{y}$ and $\hat{y}_{m_0}^{SP}$ for $m_0$ ($\mathcal{L} _{ce}$ and $\mathcal{L} _{ce}^{m_1}$ for $m_1$).
As for distillation, inspired from \cite{peng2022balanced}, we use the ground truth probabilities to monitor the performance discrepancy between two modalities:
\begin{equation}
    \begin{aligned}
        s_{k}^{m_0}=\sum_{c=1}^C{1_{c=y_k}\cdot \mathrm{soft}\max \left( \hat{y}_{m_0}^{SP}|_k \right) _c} \\
        s_{k}^{m_1}=\sum_{c=1}^C{1_{c=y_k}\cdot \mathrm{soft}\max \left( \hat{y}_{m_1}^{SP}|_k \right) _c}
        \label{eq: ground-truth prob}
    \end{aligned}
\end{equation}
\begin{equation}
    \rho _{t}^{m}=\frac{\sum\nolimits_{k\in B_t}^{}{s_{k}^{m_0}}}{\sum\nolimits_{k\in B_t}^{}{s_{k}^{m_1}}}
    \label{eq: discrepancy ratio}
\end{equation}
where $\rho _{t}^{m}$ is the discrepancy ratio. $\rho _{t}^{m}>1$ means modal $m_0$ is better than $m_1$ and vice verse. $B_t$ is a random mini-batch at time step $t$. 

With the discrepancy ratio, we use the response-based distillation loss \cite{chen2017learning, hinton2015distilling} to transfer the knowledge from the global dominant modality to the local weak modality:
\begin{equation}
\begin{split}
\mathcal{L} _{rd}=\left\{ \begin{matrix}
	\frac{1}{\left| B_t \right|}\sum\nolimits_{k\in B_t}^{}{p_{m_0,g}^{SP}|_k\log \frac{p_{m_0,g}^{SP}|_k}{p_{m_1}^{IP}|_k}}&		\rho _{t}^{m}>1\\
	\frac{1}{\left| B_t \right|}\sum\nolimits_{k\in B_t}^{}{p_{m_1,g}^{SP}|_k\log \frac{p_{m_1,g}^{SP}|_k}{p_{m_0}^{IP}|_k}}&		\rho _{t}^{m}\leqslant1\\
\end{matrix} \right. 
\end{split}
\label{eq: rd loss}
\end{equation}
where $p=\mathrm{soft}\max \left( \hat{y}/T \right) $ denotes the output probability and $T$ is the temperature. $p_{m,g}$ means the output from the global $m$ branch model. Therefore, the multimodal training loss is: $\mathcal{L} _{ce}^{}+\mathcal{L} _{ce}^{m_0}+\mathcal{L} _{ce}^{m_1}+\kappa\mathcal{L} _{rd}$, where $\kappa$ is a hyper-parameter. 

Through $\mathcal{L} _{rd}$, we transfer the knowledge from the dominant modality towards the weak modality. 
Since the distillation signal comes from global model, the local weak modality from different clients could absorb consistent knowledge that tends to avoid the strong knowledge heterogeneity caused by modality imbalance. 
%
%
$\mathcal{L} _{ce}^{m_0}$ and $\mathcal{L} _{ce}^{m_1}$ are used to enhance the information exploitation for each local modality. They are similar with the additional losses in Gradient Blending (GB) \cite{wang2020makes}, but GB needs complex control of loss weights which is not required in our method. Combining all these loss functions simultaneously enables the exploration of each modality and the integration of knowledge from dominant modality.


The updated modules except IP participate in server-client communication and aggregation, which are defined as \textbf{base modules}. 
%
%
In addition to losses above, we add a proximal term, as in \cite{li2020federated}, on base modules to regulate the optimization process which mitigates the harm from input heterogeneity:
\begin{equation}
    \mathcal{L} _{prox}=\frac{\mu}{2}\left\| \boldsymbol{\phi }_b-\boldsymbol{\phi }_{b}^{t} \right\| ^2
    \label{eq: prox term}
\end{equation}
where $\boldsymbol{\phi }_b$ denotes the parameters of base modules. Finally, we can get the total local training loss: 
\begin{equation}
    \mathcal{L} =\mathcal{L} _{ce}^{}+ \mathcal{L} _{ce}^{m_0}+\mathcal{L} _{ce}^{m_1} +\kappa\mathcal{L} _{rd}+\mu\mathcal{L} _{prox}
    \label{eq:final loss}
\end{equation}
where $\mu$ is a hyper-parameter.

\noindent\textbf{Discussion}
(1) 
IP with distillation signal allows cross-modal knowledge integration and the separated SP performs classification relying on the information of its own modality, which ensures that weak modality does not just imitate the dominant modality.
(2) The shared classifier is essential. As discussed in \cite{snell2017prototypical,li2020prototypical}, the linear classifier learns a fixed set of weights as the mean vectors for the representations in each class. Therefore, the linear shared classifier promotes consistent convergence of the two representations from the two-projector, so that the complementary and positively correlated information of two modalities can be retained.
(3) For unimodal clients, there is no multimodal data on local side for distillation, so only the modules for local modality learning are updated, as shown in \cref{fig: framework of CMI}. 
(4) Different from previous work \cite{yang2022disentangled,li2023decoupled}, our method does not require explicit and complex modality disentanglement for cross-modal knowledge transfer only on modal-shared knowledge, but achieves the knowledge infiltration between modalities with two-projector via autonomous learning. (5) The projectors in this paper are two-layer MLPs (for both SP and IP), so the size of projector is much smaller than backbone’s. Therefore, the increase in computation and communication cost is negligible. Server and clients communicate gradients, which is the same as common FL methods, so there is no additional privacy issue. Details can be seen in Appendix.
 
%


\subsection{Class-wise Temperature Adaptation}
\label{sec: cwt}
As shown in \cref{fig: motivation}, the performance gap between different modalities is remarkable and also varies at different classes. Therefore, in order to achieve full enhancement for all classes of the weak modality during cross-modal knowledge transfer, we propose the class-wise temperature adaptation mechanism. The intuition is that the weak modality should pay attention to all the teacher's logits (whether on ground truth or not) for more thorough knowledge transfer when the gap between modalities is relatively small, while it should focus more on the logit of the ground truth so as to minimize the misclassification probability expeditiously when there exists a salient performance gap between weak modality and the dominant modality. Some methods \cite{guo2021reducing, li2023curriculum} have been proposed to modulate the temperature during the student’s learning career in KD. Different from them, we monitor the discrepancy between modalities and perform class-wise temperature modulation.

First of all, we leverage the class-wise discrepancy ratio to monitor the gap between different modalities for each class on clients:
\begin{equation}
    \begin{aligned}
        s_{k,y_k}^{m_0}&=\mathrm{soft}\max \left( \hat{y}_{m_0}^{SP}|_k \right) _{y_k}
\\
s_{k,y_k}^{m_1}&=\mathrm{soft}\max \left( \hat{y}_{m_1}^{SP}|_k \right) _{y_k}
\\
\rho _{c}^{m}&=\frac{\sum\nolimits_{k\in D_i}^{}{1_{y_k=c}s_{k,y_k}^{m_0}}}{\sum\nolimits_{k\in D_i}^{}{1_{y_k=c}s_{k,y_k}^{m_1}}}
\\
\rho ^m&=\frac{1}{C}\sum_{c=1}^C{\rho _{c}^{m}}
        \label{eq: ground-truth prob}
    \end{aligned}
\end{equation}
where $\rho ^m_c$ is the discrepancy ratio for class $c$ and $\rho ^m$ is the overall discrepancy ratio.
Next, we give the adaptation scheme for temperature when $\rho ^m>1$. The corresponding scheme when $\rho ^m<1$ can be derived directly from it: 
\begin{equation}
\begin{split}
T^c=\left\{ \begin{matrix}
	T/\left( 1+\beta \log \frac{\rho _{c}^{m}}{\rho ^m} \right)&		\rho _{c}^{m}>\rho ^m\\
	T&		\rho _{c}^{m}\leqslant \rho ^m\\
\end{matrix} \right. 
\end{split}
\label{eq: class-wise temp}
\end{equation}

The response-based distillation loss for the sample with class $c$ is $p_{m_0,c}^{SP}\log \frac{p_{m_0,c}^{SP}}{p_{m_1,c}^{IP}}$ and: 
\begin{equation}
    \begin{aligned}
        p_{m_0,c}^{P}=\mathrm{soft}\max \left( \hat{y}_{m_0,c}^{SP}/T \right) \,\,
\\
p_{m_1,c}^{IP}=\mathrm{soft}\max \left( \hat{y}_{m_1,c}^{IP}/T^c \right) 
        \label{eq: class-wise prob}
    \end{aligned}
\end{equation}

It can be seen that we only modulate the temperature for the weak modality so as to make it focus on different parts of the teacher logit according to their performance gap. If the class discrepancy ratio outperforms the overall discrepancy ratio, the temperature should be smaller otherwise it remains unchanged. Overall, the pseudo-code of FedCMI is provided in \cref{alg: FedCMI}.

\begin{algorithm}[t]
\caption{FedCMI.} 
\hspace*{0.02in} {\bf Input:} 
Number of available clients $N$, dataset $D=\left\{ D_1, D_2, \cdots , D_N \right\}$, hyper-parameters $\kappa, \mu$. \\
\hspace*{0.02in} {\bf Server Executes:} 
\begin{algorithmic}[1]
\State Initialize global model $\boldsymbol{\phi }$.  
\For{each round $t=1,2,...$} 
  \State Send the base modules $\boldsymbol{\phi}_b$ to selected clients
  \For{each client $i$ \textbf{in parallel}}
    \State $\boldsymbol{\phi }_{i, b}\gets $LocalUpdate($i$, $\boldsymbol{\phi}_b$) 
  \EndFor
  \State \textbf{end for}
  \State Update global modules $\boldsymbol{\phi}_b$ by aggregating all $\boldsymbol{\phi }_{i, b}$
\EndFor
\State \textbf{end for}
\end{algorithmic}

\noindent\noindent\hspace*{0.02in} {\bf LocalUpdate($i$, $\boldsymbol{\phi}_b$):}
\begin{algorithmic}[1]
    \For{each local epoch}
        \For{batch $\left( x_{i}^{m_0},x_{i}^{m_1},y_i \right) \in D_i$}
            \State Compute the class-wise temperature by \cref{eq: class-wise temp}.
            \State Compute the training loss $\mathcal{L}$ by \cref{eq:final loss}.
            \State Update local model $\boldsymbol{\phi }_{i, b}$ according to the loss.
        \EndFor
        \State \textbf{end for}
    \EndFor
    \State \textbf{end for}
\State \Return $\boldsymbol{\phi }_{i, b}$
\end{algorithmic}
\label{alg: FedCMI}
\end{algorithm}

\section{Experiments}
\label{sec:experiment}
\subsection{Datasets and baselines}
\noindent\textbf{CREMA-D} \cite{cao2014crema} is an audio-visual dataset for emotion recognition research, consisting of both facial and vocal emotional expressions. The emotional states can be divided into 6 classes: happy, sad, angry, fear, disgust and neutral. There are 7,442 video clips in total, which are randomly divided into 6,698 samples as the training set and 744 samples as the testing set. 

\begin{table*}[t]
    \renewcommand\arraystretch{1.1}
    \begin{center}
    \setlength{\tabcolsep}{2.4mm}{
     \begin{tabular}{c|c c c c | c c c c }
        \hline
        \hline
          & \multicolumn{4}{c|}{CREMA-D} & \multicolumn{4}{c}{AVE} \\
        \hline
        Cases & case A & case B & case C & case D & case A & case B & case C & case D \\
        \hline
        MFedAvg & 52.2 ± 1.0 & 52.9 ± 0.7 & 49.8 ± 1.8 & 51.5 ± 1.0 & 56.9 ± 0.8 & 56.5 ± 1.4 & 55.9 ± 0.9 & 55.6 ± 1.0  \\
        MFedProx & 53.9 ± 1.3 & 52.4 ± 1.4 & 50.6 ± 1.1 & 53.0 ± 1.7 & 58.4 ± 1.0 & 58.7 ± 1.5 & 57.4 ± 1.3 & 56.9 ± 1.8 \\
        MFedProto & 52.8 ± 0.7 & 53.6 ± 0.9 & 53.2 ± 1.2 & 53.5 ± 0.4 & 57.2 ± 1.2 & 56.9 ± 1.7 & 56.2 ± 1.4 & 56.3 ± 1.5 \\
        \hline
        FedOGM & 57.8 ± 0.4 & 55.2 ± 0.7 & 56.3 ± 1.1 & 55.7 ± 0.8 & 60.8 ± 1.1 & 59.3 ± 1.1 & 58.9 ± 0.9 & 57.1 ± 1.0 \\
        FedPMR & 55.7 ± 0.6 & 56.1 ± 1.2 & 53.6 ± 0.9 & 54.4 ± 0.9 & 58.2 ± 0.8 & 59.5 ± 0.7 & 58.7 ± 0.7  & 57.4 ± 0.9 \\
        \hline
        FedIoT & 53.2 ± 1.1 & 54.3 ± 1.0 & 50.5 ± 0.8 & 54.7 ± 1.2 & 56.9 ± 0.5 & 58.8 ± 0.7  & 56.1 ± 1.1 & 59.0 ± 0.6 \\
        FedMSplit & 54.9 ± 0.9 & 56.9 ± 1.3 & 52.1 ± 1.0 & 55.9 ± 1.5 & 57.8 ± 0.9 & 61.2 ± 1.2 & 57.1 ± 1.0 & 60.4 ± 1.4  \\
        \hline
        \textbf{FedCMI} & \textbf{60.9 ± 1.6} & \textbf{58.9 ± 0.6} & \textbf{59.5 ± 1.3} & \textbf{57.1 ± 1.0} & \textbf{64.7 ± 1.0} & \textbf{62.8 ± 1.4} & \textbf{63.4 ± 0.9} & \textbf{61.1 ± 0.9} \\
        \hline
        \hline
        \end{tabular} 
        \caption{Comparison of FedCMI with baselines on two audio-visual datasets: CREMA-D and AVE. FedCMI achieves significant improvement in all settings.
    }
    \label{tab: Accuracy audio-visual}
    \vspace{-10pt}
    }
    \end{center}
\end{table*}

\noindent\textbf{AVE} \cite{tian2018audio} is an audio-visual video dataset for event localization. There are a total of 28 event classes and 4,143 10-second video clips with both auditory and visual tracks as well as second-level annotations which are collected from YouTube. In our experiments, we follow the preprocessing as in \cite{fan2023pmr}, extracting the frames from event-localized video segments and capturing the audio clips within the same segments and further constructing a labeled multimodal classification dataset. 

\noindent\textbf{CrisisMMD} \cite{alam2018crisismmd} is a image-text crisis dataset, which consists of annotated image-tweet pairs collected using event-specific keywords and hashtags during seven natural disasters in 2017. We use its classification task of Informative vs. Not Informative as discribed in \cite{abavisani2020multimodal}. There are 9,601 label-consistent samples for training and 1,534 for testing.

\noindent\textbf{Baselines.} We compare FedCMI with three categories of baselines: (1) unimodal FL frameworks extended to multimodal scenario: MFedAvg, MFedProx and MFedProto, derived from FedAvg \cite{mcmahan2017communication}, FedProx \cite{li2020federated} and FedProto \cite{tan2022fedproto} respectively. (2) FedOGM and FedPMR, two MFL methods that combine FedAvg with OGM-GE \cite{peng2022balanced} and PMR \cite{fan2023pmr}, the solutions for modality imbalance. (3) Existed MFL methods designed for input heterogeneity: FedIoT \cite{zhao2022multimodal} and FedMSplit \cite{chen2022fedmsplit}.

\subsection{Implementation Details}
We consider the stimulation of both statistical and modality heterogeneity in MFL. We use Dirichlet distribution ($\alpha=5,3,3$ for CREMA-D, AVE and CrisisMMD respectively) for non-IID data partitions \cite{hsu2019measuring}. For modality heterogeneity, we select 50\% of clients to preserve both modalities, and the remaining clients only retain one modality data in random. Depending on whether it is heterogeneous or not, we design four experimental cases: case A: IID multimodal data among participating clients without modality heterogeneity; case B: IID multimodal data with modality heterogeneity; case C: non-IID multimodal data without modality heterogeneity; case D: non-IID multimodal data with modality heterogeneity.

For CREMA-D and AVE, we use ResNet18 \cite{he2016deep} as the basic encoder for both audio and visual modalities as in \cite{peng2022balanced}. For CrisisMMD, we use DenseNet121 \cite{huang2017densely} as image encoder and Wikipedia pre-trained BERT \cite{devlin2018bert} as text encoder. 
For projectors, we use a two-layer MLP by default. The fusion method is concatenation \cite{owens2018audio} for three datasets unless otherwise specified. Data preprocessing of CREMA-D and AVE follows \cite{fan2023pmr}. There are a total of 20 clients for CREMA-D and AVE, 50 clients for CrisisMMD. In each round, 5 clients are selected for CREMA-D and AVE, and the number for CrisisMMD is 10. The number of local training epochs is set to 5. 
The optimizer is SGD for all datasets, where learning rate is 1e-3 for CREMA-D and AVE, 2e-3 for CrisisMMD.
The two hyperparameters $\kappa$ and $\mu$ is set to 1-3 and 1 respectively according to the experimental settings. 

\begin{table}[t]
    \renewcommand\arraystretch{1.1}
    \begin{center}
    \setlength{\tabcolsep}{1.5mm}{
            \begin{tabular}{c|c c c c}
        \hline
        \hline
          & \multicolumn{4}{c}{CrisisMMD} \\
        \hline
        Cases & case A & case B & case C & case D \\
        \hline
        MFedAvg & 86.0±0.6 & 86.2±0.7 & 85.7±1.8 & 84.7±2.0 \\
        MFedProx & 86.3±0.4 & 85.4±0.2 & 86.0±0.3 & 85.1±0.5  \\
        MFedProto & 86.9±0.6 & 86.5±0.6 & 87.0±0.9 & 85.4±1.1  \\
        \hline
        FedOGM & 86.7±0.3 & 85.9±0.7 & 86.6±1.0 & 84.8±0.4  \\
        FedPMR & 87.6±0.4 & 86.6±0.3 & 87.5±0.6 & 86.7±0.5   \\
        \hline
        FedIoT & 86.2±0.8 & 86.3±0.7 & 85.4±0.9 & 83.8±1.3   \\
        FedMSplit & 86.7±0.6 & \textbf{87.8±0.6} & 86.9±0.8 & 87.0±0.7   \\
        \hline
        \textbf{FedCMI} & \textbf{88.5±0.5} & 87.3±0.3 & \textbf{88.0±0.6} & \textbf{87.2±0.4}  \\
        \hline
        \hline
        \end{tabular}
        \caption{Comparison of FedCMI with other baselines on the image-text dataset, CrisisMMD. FedCMI achieves the best performance in almost all settings.}
    \label{tab: Accuracy image-text}
    }
    \end{center}
    \vspace{-14pt}
\end{table}

\subsection{Main Results}
\label{sec: main results}
\noindent\textbf{Performance comparison with baselines.} \cref{tab: Accuracy audio-visual} presents the performance of baselines and our method in audio-visual tasks with the four different cases. It can be seen that our FedCMI achieves noticeable performance improvement overall baselines. 
In comparison to MFedAvg, MFedProx and MFedProto, 
FedCMI shows superiority on the global model performance (up to 7.0\% and 6.3\% accuracy improvement respectively on CREMA-D and AVE). These baselines only consider the statistical heterogeneity in FL and ignore the complementary information of multimodal data, which restricts their performance.
Compared to FedOGM and FedPMR,
FedCMI also achieves superior performance (up to 3.2\% and 4.5\% accuracy improvement respectively on CREMA-D and AVE), indicating the necessity of considering modality imbalance in distributed scenarios and the importance of knowledge transfer from the dominant modality.
%
By contrast with FedIoT and FedMSplit, FedCMI still exhibits particularly better performance since these baselines ignore the severe heterogeneous modality inhibition. Besides, these two baselines could perform better on case B, D than on case A, C because they are designed for modality heterogeneity.
To prove our method’s effectiveness beyond audio-visual modality, we further present experimental results on image-text modality as shown in \cref{tab: Accuracy image-text}. FedCMI also achieves appreciable improvement on all settings. 
Besides, there are a total of 50 clients for CrisisMMD, showing the applicability of our method in large-scale scenarios.

\begin{figure}[t]
    \centering
    \includegraphics[width=0.8\linewidth]{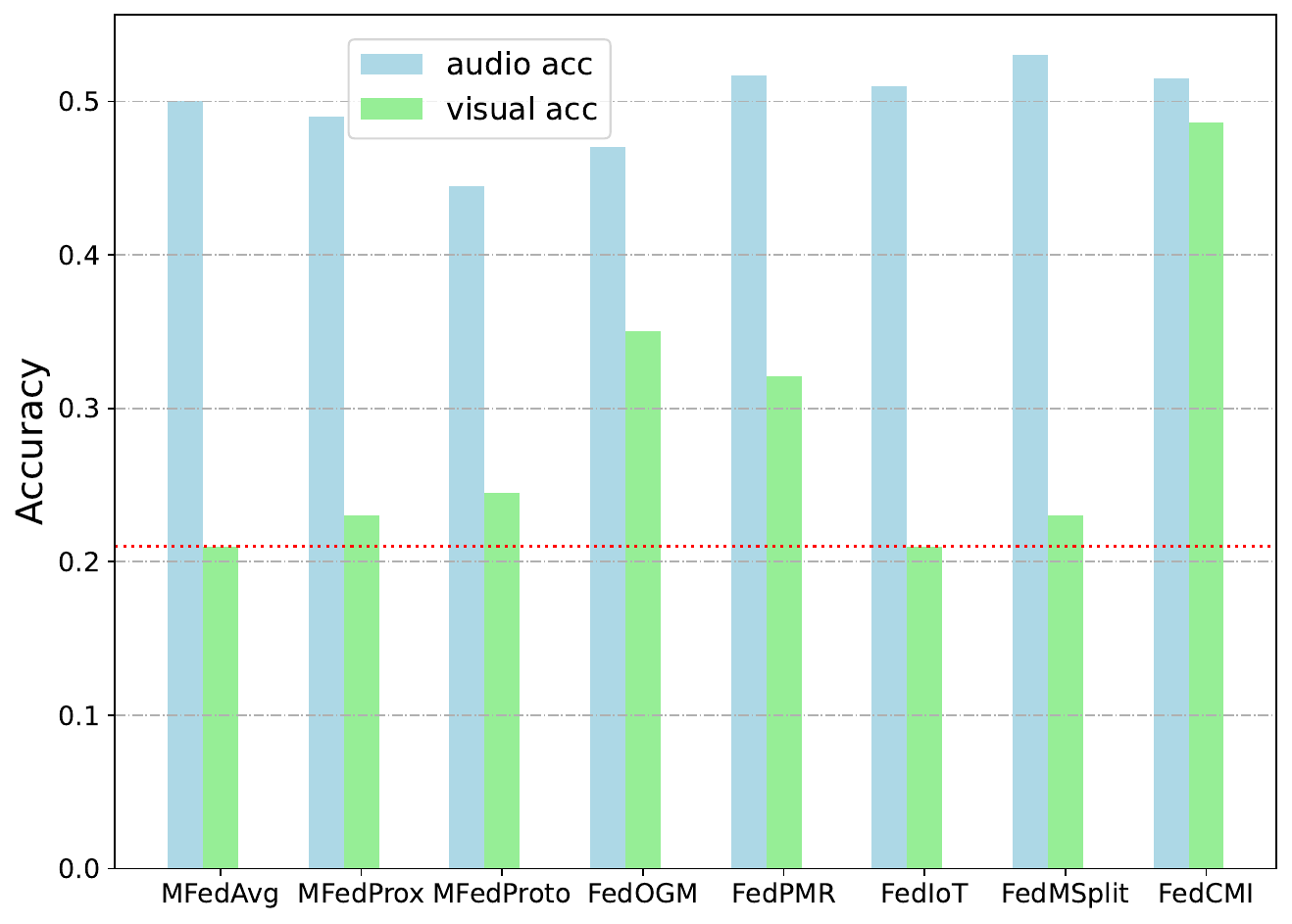}
    \caption{The performance of uni-modality in MFL on CREMA-D dataset in case A. The visual modality is extremely inhibited in baselines. Our method not only effectively improves the performance of visual modal, but also makes improvements on the overall performance.}
    \label{fig: baseline modality accuracy}
\end{figure}
\begin{figure}[t]
    \centering
    \includegraphics[width=0.8\linewidth]{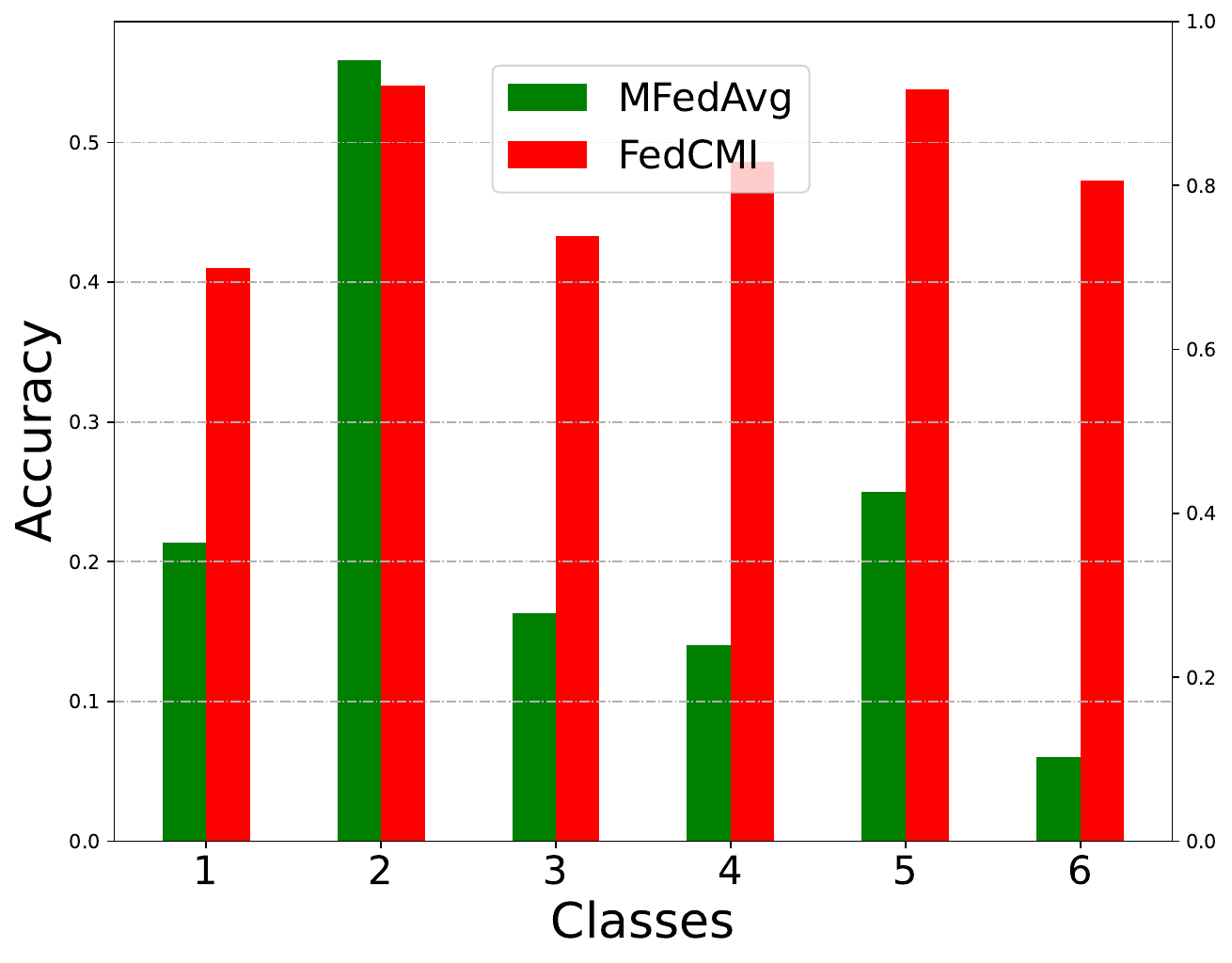}
    \caption{The class-wise performance of visual modality from the global model on CREMA-D under case A. MFedAvg leads to particularly biased knowledge and FedCMI learns balanced performance across classes.}
    \label{fig: FedCMI class-wise accuracy}
\end{figure}

\noindent\textbf{Unimodal performance comparison.} The main goal of FedCMI is to release the ability of the inhibited modality to realize full information exploitation in MFL. Therefore, we exhibit the unimodal performance of the global model from FedCMI and all the baselines, as shown in \cref{fig: baseline modality accuracy}. 
It's clear that our FedCMI can significantly improve the performance of the weak modality (visual here) by transferring knowledge from the global dominant modality (minimal performance gap between two modalities) without compromising its performance on audio modality. 
Although FedOGM and FedPMR can alleviate modality imbalance to some extent, they are struggling in MFL as they can only use local data to modulate the learning process while ignores the serious inhibition heterogeneity on the weak modality. 
Moreover, we find that some methods could improve the performance of multimodal networks, while they sacrifice the performance of the dominant modality (e.g., MfedProx, MFedProto and FedOGM). FedIoT and FedMSplit improve the performance of the audio modality through a better aggregation strategy which leads to improvement compared with MFedAvg.
%

To show that our method does alleviate the strong bias of learned knowledge on different classes, we illustrate the performance of each class on CREMA-D in case A, as shown in \cref{fig: FedCMI class-wise accuracy}. It is apparent from this figure that in MFedAvg, there exists a strong bias of the visual modal network (e.g., the accuracy on class 2 is almost 10 times the accuracy on class 6 ).
In contrast, our FedCMI achieves more balanced performance over all classes as the distilled knowledge from the global dominant modality and the class-wise temperature adaptation can reduce the gap between different classes.

\begin{figure}[t]
    \centering
    \includegraphics[width=0.9\linewidth]{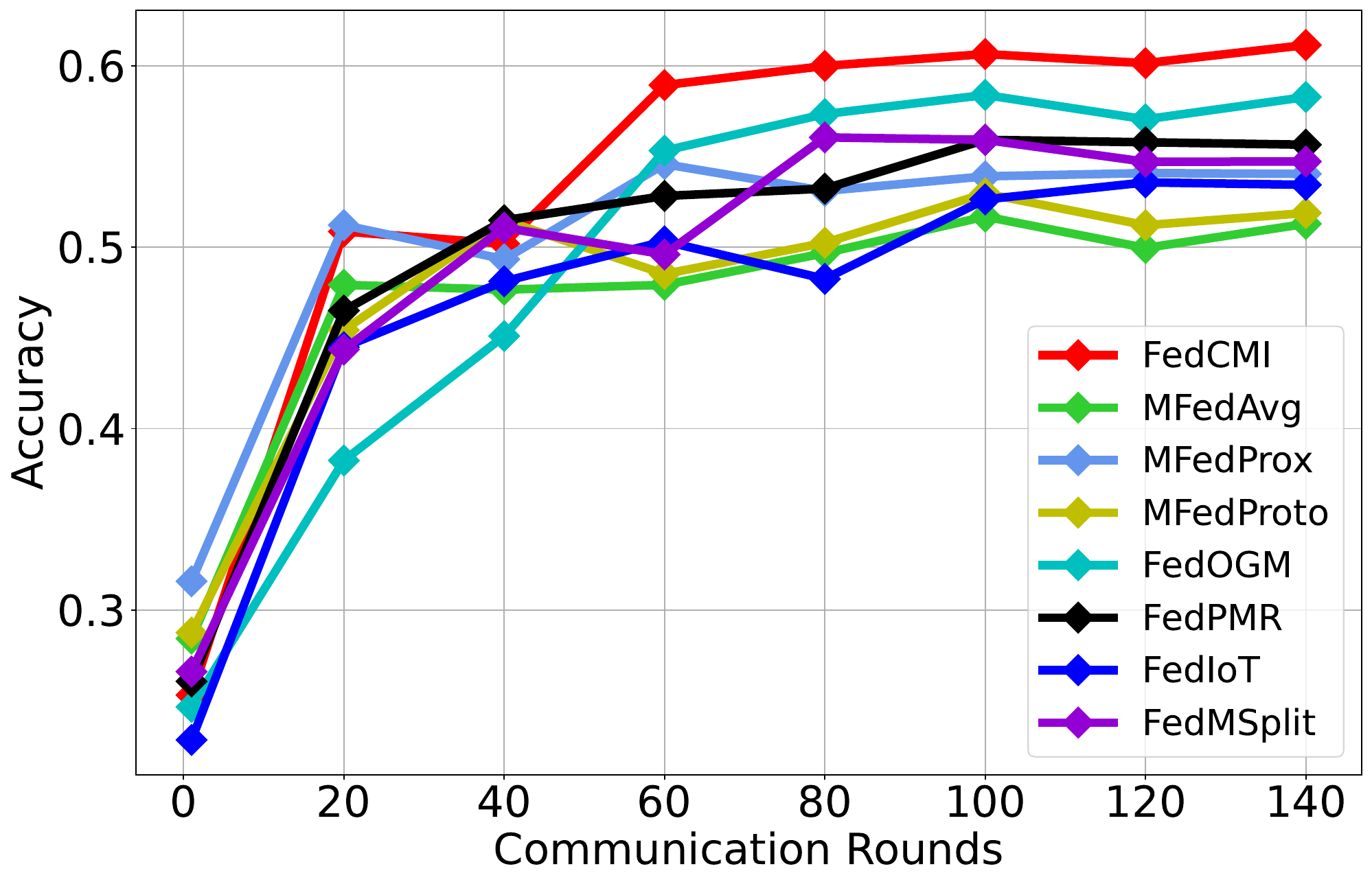}
    \caption{Test accuracy versus number of communication rounds for all baselines and our FedCMI. Experiments on CREMA-D under the ``full/IID" setting. FedCMI converges fast and consistently outperforms strong competitors.}
    \label{fig: training curves}
    \vspace{-1.0em}
\end{figure}

\noindent\textbf{Communication rounds.} Although our framework has the two-projector modules, the extra training and communication cost is negligible as the projector is only a tiny two-layer MLP and the infiltration projector does not participate in server-client communication. Therefore, the communication overhead is mainly related to the communication round. As shown in \cref{fig: training curves}, our proposed FedCMI framework not only achieves the best performance, but also converges faster to the desired accuracy. 
Although FedOGM and FedPMR could converge to better results than MFedAvg, they may slow down the learning of dominant modality or just depend on local information, which results in the requirement of more MFL rounds for convergence.

\begin{table}[t]
  \centering
  \begin{tabular}{c  c  c  c | c | c}
  \hline
  \hline
    \makebox[0.04\textwidth][c]{TP} & \makebox[0.04\textwidth][c]{RD} & \makebox[0.04\textwidth][c]{CTA} & \makebox[0.04\textwidth][c]{PT} & \makebox[0.072\textwidth][c]{case A} & \makebox[0.072\textwidth][c]{case C} \\
    \hline
    \ding{55} & \ding{55} & \ding{55}  & \ding{55}  & 52.2 & 51.5 \\
    \hline
    \ding{52} &   &   &  & 54.4 & 54.0 \\ 

    \ding{52} & \ding{52} &   &   & 58.9 & 55.4 \\

    \ding{52} & \ding{52} & \ding{52} &   & 60.2 & 55.9 \\
    UP & \ding{52} & \ding{52} & \ding{52} & 56.7 & 54.3 \\
    \ding{52} & \ding{52} & \ding{52} & \ding{52} & \textbf{60.9} & \textbf{57.1} \\
    \hline
  \hline
  \end{tabular}
  \caption{Results on ablation studies on CREMA-D under case A and D. \textit{TP}: two-projector module. \textit{RD}: relation-based distillation. \textit{CTA}: class-wise temperature adaptation. \textit{PT}: proximal term. \textit{UP}: uni-projector module. Each component is essential in our framework.}
  \label{tab: 1.0/IID}
\end{table}





\subsection{Ablation studies}
\noindent\textbf{Effect of each FedCMI component.} \cref{tab: 1.0/IID} studies the effect of each component in our FedCMI. \textit{TP} denotes the two-projector module with its cross-entropy loss for each modality. \textit{RD} is the response-based distillation loss. \textit{CTA} means class-wise temperature adaptation method. \textit{PT} adopts proximal term on the base modules. 
\textit{UP} here represents replacing two-projector with a uni-projector, the output logit of which is used both for self-enhancement of cross-entropy loss and cross-modal knowledge transfer of distillation loss.
Our FedCMI degenerates into MFedAvg if all these components are missing.
The results are conducted on CREMA-D under case A and D respectively.
As shown in \cref{tab: 1.0/IID}, applying TP surpasses MFedAvg by an obvious margin, as the weak modality can be promoted through TP via enhancement from local information. 
What is striking about applying TP is that it also exceeds MFedProx and MFedProto by a small margin (e.g., 53.9\% and 52.8\% v.s. 54.4\% under case A).
RD is based on the two-projector module. Applying RD makes an impressive improvement compared with the vanilla strategy and also FedOGM and FedPMR (57.8\% and 55.7\% v.s. 58.9\% in case A), demonstrating the success of transferring knowledge of global dominant modality to the weak modality. The combination of TP and RD alleviates modality imbalance through local and global information simultaneously.
Combing CTA can further alleviate the biased knowledge learned by the weak modality, leading to less heterogeneity across clients and further improvement. Adding proximal term on base modules aims to alleviate the harm from input heterogeneity as in traditional FL. Although PT does not make impressive improvement in the case A, it works well under the non-IID setting (case D, 55.9\% v.s. 57.1\%), which shows that our method does not conflict with the previous methods for data heterogeneity, and can complement and promote each other. The three other components also demonstrate their effectiveness in non-IID settings. 
Compare the results of the last two rows, it is apparent that using uni-projector for both self-enhancement and integrating the knowledge from the dominant modality is not enough since strong imitation may impede exploration of the modality itself. 
%

\noindent\textbf{Effectiveness on other fusion methods.} \cref{tab: fusion methods} shows the results with different fusion methods on CREMA-D under case A. Here, we choose other three common fusion methods: summation, film \cite{perez2018film} and gated \cite{kiela2018efficient}. Summation and the above concatenation are two simple fusion methods, while in film and gated, features between modalities are fused in more complicated ways. 
FedCMI exhibits superior performance in all fusion methods, demonstrating the great generalization of our method on different structures.

  

\begin{table}
  \centering
  \begin{tabular}{c | c | c | c}
  \hline
  \hline
    \makebox[0.072\textwidth][c]{Method} & \makebox[0.072\textwidth][c]{Sum} & \makebox[0.072\textwidth][c]{Film} & \makebox[0.072\textwidth][c]{Gated} \\
    \hline
    MFedAvg & 51.8 & 53.1  & 50.9 \\
    MFedProx & 52.6 & 51.9 & 54.3 \\
    MFedProto & 52.5 & 53.7 & 53.2\\
    FedOGM & 57.3 & 58.0 & 57.7 \\
    FedPMR & 55.3 & 54.4 & 54.2 \\
    FedIoT & 52.7 & 53.6 & 53.1 \\
    FedMSplit & 55.1 & 56.2 & 54.3 \\
    \hline
    \textbf{FedCMI} & \textbf{59.6} & \textbf{61.4} & \textbf{60.6} \\
    \hline
  \hline
  \end{tabular}
  \caption{The comparison results of FedCMI with all baselines on CREMA-D under the ``full/IID" setting with various fusion methods. FedCMI gets better performance compared with baselines on all scenarios by a large margin.}
  \label{tab: fusion methods}
\end{table}
\section{Conclusion}
\label{sec:conclusion}
Jointly considering modality imbalance and input heterogeneity makes 
MFL encounter severe heterogeneous modality inhibition over weak modality, which can potentially undermine the generalization capability of the produced global model. In this paper, we propose a cross-modal infiltration federated learning (FedCMI) framework to balance the modality-level performance for distributed users by transferring the global dominant knowledge to the weak modality and maintaining its own feature exploitation. The two-projector and shared classifier ensure that knowledge from different modalities can co-exist in a single modal network. We also design a class-wise temperature adaptation scheme to mitigate the remarkable performance divergence between different classes. It is demonstrated that our method achieves considerable performance improvement on the two audio-visual datasets and a image-text dataset under various settings. The ablation study also shows FedCMI's effectiveness for different fusion structures.
{
    \small
    \bibliographystyle{ieeenat_fullname}
    \bibliography{main}

\begin{thebibliography}{43}
\providecommand{\natexlab}[1]{#1}
\providecommand{\url}[1]{\texttt{#1}}
\expandafter\ifx\csname urlstyle\endcsname\relax
  \providecommand{\doi}[1]{doi: #1}\else
  \providecommand{\doi}{doi: \begingroup \urlstyle{rm}\Url}\fi

\bibitem[Abavisani et~al.(2020)Abavisani, Wu, Hu, Tetreault, and Jaimes]{abavisani2020multimodal}
Mahdi Abavisani, Liwei Wu, Shengli Hu, Joel~R Tetreault, and Alejandro Jaimes.
\newblock Multimodal categorization of crisis events in social media. in 2020 ieee.
\newblock In \emph{CVF Conference on Computer Vision and Pattern Recognition, CVPR}, pages 13--19, 2020.

\bibitem[Alam et~al.(2018)Alam, Ofli, and Imran]{alam2018crisismmd}
Firoj Alam, Ferda Ofli, and Muhammad Imran.
\newblock Crisismmd: Multimodal twitter datasets from natural disasters.
\newblock In \emph{Proceedings of the international AAAI conference on web and social media}, 2018.

\bibitem[Byrd and Polychroniadou(2020)]{byrd2020differentially}
David Byrd and Antigoni Polychroniadou.
\newblock Differentially private secure multi-party computation for federated learning in financial applications.
\newblock In \emph{Proceedings of the First ACM International Conference on AI in Finance}, pages 1--9, 2020.

\bibitem[Cao et~al.(2014)Cao, Cooper, Keutmann, Gur, Nenkova, and Verma]{cao2014crema}
Houwei Cao, David~G Cooper, Michael~K Keutmann, Ruben~C Gur, Ani Nenkova, and Ragini Verma.
\newblock Crema-d: Crowd-sourced emotional multimodal actors dataset.
\newblock \emph{IEEE transactions on affective computing}, 5\penalty0 (4):\penalty0 377--390, 2014.

\bibitem[Chen et~al.(2017)Chen, Choi, Yu, Han, and Chandraker]{chen2017learning}
Guobin Chen, Wongun Choi, Xiang Yu, Tony Han, and Manmohan Chandraker.
\newblock Learning efficient object detection models with knowledge distillation.
\newblock \emph{Advances in neural information processing systems}, 30, 2017.

\bibitem[Chen and Zhang(2022)]{chen2022fedmsplit}
Jiayi Chen and Aidong Zhang.
\newblock Fedmsplit: Correlation-adaptive federated multi-task learning across multimodal split networks.
\newblock In \emph{Proceedings of the 28th ACM SIGKDD Conference on Knowledge Discovery and Data Mining}, pages 87--96, 2022.

\bibitem[Chen and Li(2022)]{chen2022towards}
Sijia Chen and Baochun Li.
\newblock Towards optimal multi-modal federated learning on non-iid data with hierarchical gradient blending.
\newblock In \emph{IEEE INFOCOM 2022-IEEE Conference on Computer Communications}, pages 1469--1478. IEEE, 2022.

\bibitem[Chen et~al.(2020)Chen, Kornblith, Norouzi, and Hinton]{chen2020simple}
Ting Chen, Simon Kornblith, Mohammad Norouzi, and Geoffrey Hinton.
\newblock A simple framework for contrastive learning of visual representations.
\newblock In \emph{International conference on machine learning}, pages 1597--1607. PMLR, 2020.

\bibitem[Dai et~al.(2023)Dai, Chen, Li, Heinecke, Sun, and Xu]{dai2023tackling}
Yutong Dai, Zeyuan Chen, Junnan Li, Shelby Heinecke, Lichao Sun, and Ran Xu.
\newblock Tackling data heterogeneity in federated learning with class prototypes.
\newblock In \emph{Proceedings of the AAAI Conference on Artificial Intelligence}, pages 7314--7322, 2023.

\bibitem[Devlin et~al.(2018)Devlin, Chang, Lee, and Toutanova]{devlin2018bert}
Jacob Devlin, Ming-Wei Chang, Kenton Lee, and Kristina Toutanova.
\newblock Bert: Pre-training of deep bidirectional transformers for language understanding.
\newblock \emph{arXiv preprint arXiv:1810.04805}, 2018.

\bibitem[Fan et~al.(2023)Fan, Xu, Wang, Wang, and Guo]{fan2023pmr}
Yunfeng Fan, Wenchao Xu, Haozhao Wang, Junxiao Wang, and Song Guo.
\newblock Pmr: Prototypical modal rebalance for multimodal learning.
\newblock In \emph{Proceedings of the IEEE/CVF Conference on Computer Vision and Pattern Recognition}, pages 20029--20038, 2023.

\bibitem[Guo(2021)]{guo2021reducing}
Jia Guo.
\newblock Reducing the teacher-student gap via adaptive temperatures.
\newblock 2021.

\bibitem[He et~al.(2016)He, Zhang, Ren, and Sun]{he2016deep}
Kaiming He, Xiangyu Zhang, Shaoqing Ren, and Jian Sun.
\newblock Deep residual learning for image recognition.
\newblock In \emph{Proceedings of the IEEE conference on computer vision and pattern recognition}, pages 770--778, 2016.

\bibitem[Hinton et~al.(2015)Hinton, Vinyals, and Dean]{hinton2015distilling}
Geoffrey Hinton, Oriol Vinyals, and Jeff Dean.
\newblock Distilling the knowledge in a neural network.
\newblock \emph{arXiv preprint arXiv:1503.02531}, 2015.

\bibitem[Hsu et~al.(2019)Hsu, Qi, and Brown]{hsu2019measuring}
Tzu-Ming~Harry Hsu, Hang Qi, and Matthew Brown.
\newblock Measuring the effects of non-identical data distribution for federated visual classification.
\newblock \emph{arXiv preprint arXiv:1909.06335}, 2019.

\bibitem[Huang et~al.(2017)Huang, Liu, Van Der~Maaten, and Weinberger]{huang2017densely}
Gao Huang, Zhuang Liu, Laurens Van Der~Maaten, and Kilian~Q Weinberger.
\newblock Densely connected convolutional networks.
\newblock In \emph{Proceedings of the IEEE conference on computer vision and pattern recognition}, pages 4700--4708, 2017.

\bibitem[Kiela et~al.(2018)Kiela, Grave, Joulin, and Mikolov]{kiela2018efficient}
Douwe Kiela, Edouard Grave, Armand Joulin, and Tomas Mikolov.
\newblock Efficient large-scale multi-modal classification.
\newblock In \emph{Proceedings of the AAAI conference on artificial intelligence}, 2018.

\bibitem[Li et~al.(2020{\natexlab{a}})Li, Zhou, Xiong, and Hoi]{li2020prototypical}
Junnan Li, Pan Zhou, Caiming Xiong, and Steven~CH Hoi.
\newblock Prototypical contrastive learning of unsupervised representations.
\newblock \emph{arXiv preprint arXiv:2005.04966}, 2020{\natexlab{a}}.

\bibitem[Li et~al.(2020{\natexlab{b}})Li, Sahu, Zaheer, Sanjabi, Talwalkar, and Smith]{li2020federated}
Tian Li, Anit~Kumar Sahu, Manzil Zaheer, Maziar Sanjabi, Ameet Talwalkar, and Virginia Smith.
\newblock Federated optimization in heterogeneous networks.
\newblock \emph{Proceedings of Machine learning and systems}, 2:\penalty0 429--450, 2020{\natexlab{b}}.

\bibitem[Li et~al.(2023{\natexlab{a}})Li, Wang, and Cui]{li2023decoupled}
Yong Li, Yuanzhi Wang, and Zhen Cui.
\newblock Decoupled multimodal distilling for emotion recognition.
\newblock In \emph{Proceedings of the IEEE/CVF Conference on Computer Vision and Pattern Recognition}, pages 6631--6640, 2023{\natexlab{a}}.

\bibitem[Li et~al.(2023{\natexlab{b}})Li, Li, Yang, Zhao, Song, Luo, Li, and Yang]{li2023curriculum}
Zheng Li, Xiang Li, Lingfeng Yang, Borui Zhao, Renjie Song, Lei Luo, Jun Li, and Jian Yang.
\newblock Curriculum temperature for knowledge distillation.
\newblock In \emph{Proceedings of the AAAI Conference on Artificial Intelligence}, pages 1504--1512, 2023{\natexlab{b}}.

\bibitem[Liu et~al.(2020)Liu, Wu, Ge, Fan, and Zou]{liu2020federated}
Fenglin Liu, Xian Wu, Shen Ge, Wei Fan, and Yuexian Zou.
\newblock Federated learning for vision-and-language grounding problems.
\newblock In \emph{Proceedings of the AAAI Conference on Artificial Intelligence}, pages 11572--11579, 2020.

\bibitem[McMahan et~al.(2017)McMahan, Moore, Ramage, Hampson, and y~Arcas]{mcmahan2017communication}
Brendan McMahan, Eider Moore, Daniel Ramage, Seth Hampson, and Blaise~Aguera y Arcas.
\newblock Communication-efficient learning of deep networks from decentralized data.
\newblock In \emph{Artificial intelligence and statistics}, pages 1273--1282. PMLR, 2017.

\bibitem[Owens and Efros(2018)]{owens2018audio}
Andrew Owens and Alexei~A Efros.
\newblock Audio-visual scene analysis with self-supervised multisensory features.
\newblock In \emph{Proceedings of the European conference on computer vision (ECCV)}, pages 631--648, 2018.

\bibitem[Peng et~al.(2022)Peng, Wei, Deng, Wang, and Hu]{peng2022balanced}
Xiaokang Peng, Yake Wei, Andong Deng, Dong Wang, and Di Hu.
\newblock Balanced multimodal learning via on-the-fly gradient modulation.
\newblock In \emph{Proceedings of the IEEE/CVF Conference on Computer Vision and Pattern Recognition}, pages 8238--8247, 2022.

\bibitem[Perez et~al.(2018)Perez, Strub, De~Vries, Dumoulin, and Courville]{perez2018film}
Ethan Perez, Florian Strub, Harm De~Vries, Vincent Dumoulin, and Aaron Courville.
\newblock Film: Visual reasoning with a general conditioning layer.
\newblock In \emph{Proceedings of the AAAI conference on artificial intelligence}, 2018.

\bibitem[Pfitzner et~al.(2021)Pfitzner, Steckhan, and Arnrich]{pfitzner2021federated}
Bjarne Pfitzner, Nico Steckhan, and Bert Arnrich.
\newblock Federated learning in a medical context: A systematic literature review.
\newblock \emph{ACM Transactions on Internet Technology (TOIT)}, 21\penalty0 (2):\penalty0 1--31, 2021.

\bibitem[Snell et~al.(2017)Snell, Swersky, and Zemel]{snell2017prototypical}
Jake Snell, Kevin Swersky, and Richard Zemel.
\newblock Prototypical networks for few-shot learning.
\newblock \emph{Advances in neural information processing systems}, 30, 2017.

\bibitem[Tan et~al.(2022)Tan, Long, Liu, Zhou, Lu, Jiang, and Zhang]{tan2022fedproto}
Yue Tan, Guodong Long, Lu Liu, Tianyi Zhou, Qinghua Lu, Jing Jiang, and Chengqi Zhang.
\newblock Fedproto: Federated prototype learning across heterogeneous clients.
\newblock In \emph{Proceedings of the AAAI Conference on Artificial Intelligence}, pages 8432--8440, 2022.

\bibitem[Tian et~al.(2022)Tian, Wang, Xu, Cao, Shen, and Shen]{tian2022multimodal}
Jialin Tian, Kai Wang, Xing Xu, Zuo Cao, Fumin Shen, and Heng~Tao Shen.
\newblock Multimodal disentanglement variational autoencoders for zero-shot cross-modal retrieval.
\newblock In \emph{Proceedings of the 45th International ACM SIGIR Conference on Research and Development in Information Retrieval}, pages 960--969, 2022.

\bibitem[Tian et~al.(2018)Tian, Shi, Li, Duan, and Xu]{tian2018audio}
Yapeng Tian, Jing Shi, Bochen Li, Zhiyao Duan, and Chenliang Xu.
\newblock Audio-visual event localization in unconstrained videos.
\newblock In \emph{Proceedings of the European conference on computer vision (ECCV)}, pages 247--263, 2018.

\bibitem[Wang et~al.(2020)Wang, Tran, and Feiszli]{wang2020makes}
Weiyao Wang, Du Tran, and Matt Feiszli.
\newblock What makes training multi-modal classification networks hard?
\newblock In \emph{Proceedings of the IEEE/CVF Conference on Computer Vision and Pattern Recognition}, pages 12695--12705, 2020.

\bibitem[Wang et~al.(2021)Wang, Chen, and Zhu]{wang2021multimodal}
Xin Wang, Hong Chen, and Wenwu Zhu.
\newblock Multimodal disentangled representation for recommendation.
\newblock In \emph{2021 IEEE International Conference on Multimedia and Expo (ICME)}, pages 1--6. IEEE, 2021.

\bibitem[Wu et~al.(2022)Wu, Jastrzebski, Cho, and Geras]{wu2022characterizing}
Nan Wu, Stanislaw Jastrzebski, Kyunghyun Cho, and Krzysztof~J Geras.
\newblock Characterizing and overcoming the greedy nature of learning in multi-modal deep neural networks.
\newblock In \emph{International Conference on Machine Learning}, pages 24043--24055. PMLR, 2022.

\bibitem[Xiao et~al.(2020)Xiao, Lee, Grauman, Malik, and Feichtenhofer]{xiao2020audiovisual}
Fanyi Xiao, Yong~Jae Lee, Kristen Grauman, Jitendra Malik, and Christoph Feichtenhofer.
\newblock Audiovisual slowfast networks for video recognition.
\newblock \emph{arXiv preprint arXiv:2001.08740}, 2020.

\bibitem[Xiong et~al.(2022)Xiong, Yang, Qi, and Xu]{xiong2022unified}
Baochen Xiong, Xiaoshan Yang, Fan Qi, and Changsheng Xu.
\newblock A unified framework for multi-modal federated learning.
\newblock \emph{Neurocomputing}, 480:\penalty0 110--118, 2022.

\bibitem[Xue et~al.(2022)Xue, Gao, Ren, and Zhao]{xue2022modality}
Zihui Xue, Zhengqi Gao, Sucheng Ren, and Hang Zhao.
\newblock The modality focusing hypothesis: Towards understanding crossmodal knowledge distillation.
\newblock In \emph{The Eleventh International Conference on Learning Representations}, 2022.

\bibitem[Yang et~al.(2022)Yang, Huang, Kuang, Du, and Zhang]{yang2022disentangled}
Dingkang Yang, Shuai Huang, Haopeng Kuang, Yangtao Du, and Lihua Zhang.
\newblock Disentangled representation learning for multimodal emotion recognition.
\newblock In \emph{Proceedings of the 30th ACM International Conference on Multimedia}, pages 1642--1651, 2022.

\bibitem[Yao and Mihalcea(2022)]{yao2022modality}
Yiqun Yao and Rada Mihalcea.
\newblock Modality-specific learning rates for effective multimodal additive late-fusion.
\newblock In \emph{Findings of the Association for Computational Linguistics: ACL 2022}, pages 1824--1834, 2022.

\bibitem[Yu et~al.(2023)Yu, Liu, Wang, Xu, and Liu]{yu2023multimodal}
Qiying Yu, Yang Liu, Yimu Wang, Ke Xu, and Jingjing Liu.
\newblock Multimodal federated learning via contrastive representation ensemble.
\newblock \emph{arXiv preprint arXiv:2302.08888}, 2023.

\bibitem[Zhao et~al.(2022)Zhao, Barnaghi, and Haddadi]{zhao2022multimodal}
Yuchen Zhao, Payam Barnaghi, and Hamed Haddadi.
\newblock Multimodal federated learning on iot data.
\newblock In \emph{2022 IEEE/ACM Seventh International Conference on Internet-of-Things Design and Implementation (IoTDI)}, pages 43--54. IEEE, 2022.

\bibitem[Zheng et~al.(2023)Zheng, Li, Chen, Wang, and Luo]{zheng2023autofed}
Tianyue Zheng, Ang Li, Zhe Chen, Hongbo Wang, and Jun Luo.
\newblock Autofed: Heterogeneity-aware federated multimodal learning for robust autonomous driving.
\newblock \emph{arXiv preprint arXiv:2302.08646}, 2023.

\bibitem[Zheng et~al.(2022)Zheng, Zhou, Sun, Wang, Liu, and Li]{zheng2022applications}
Zhaohua Zheng, Yize Zhou, Yilong Sun, Zhang Wang, Boyi Liu, and Keqiu Li.
\newblock Applications of federated learning in smart cities: recent advances, taxonomy, and open challenges.
\newblock \emph{Connection Science}, 34\penalty0 (1):\penalty0 1--28, 2022.

\end{thebibliography}
}


\end{document}